\documentclass[10pt,conference]{IEEEtran}
\IEEEoverridecommandlockouts
\usepackage{multirow}
\usepackage{cite}
\usepackage{amsmath,amssymb,amsfonts}
\usepackage{algorithmic}
\usepackage{graphicx}
\usepackage{textcomp}
\usepackage{xcolor}
\usepackage{multirow}
\usepackage{subcaption}
\usepackage{enumitem}
\usepackage{algorithm}
\usepackage[ruled,linesnumbered,algo2e]{algorithm2e}
\usepackage{tcolorbox}
\usepackage{listings}
\usepackage{booktabs}
\usepackage{graphicx} 
\usepackage{bm}
\usepackage{xspace}
\usepackage{caption}
\usepackage{url}
\usepackage{amsmath,amssymb,amsfonts}
\usepackage{textcomp}
\usepackage{xcolor}
\usepackage{cite}
\usepackage{tcolorbox}
\usepackage{color,xcolor,colortbl}
\usepackage{graphicx}
\usepackage{mathtools}

\usepackage[ruled,linesnumbered,algo2e]{algorithm2e}
\def\BibTeX{{\rm B\kern-.05em{\sc i\kern-.025em b}\kern-.08em
    T\kern-.1667em\lower.7ex\hbox{E}\kern-.125emX}}

\DontPrintSemicolon
\SetAlgoNlRelativeSize{0}
\SetAlgoNlRelativeSize{-1}

\definecolor{mygray}{gray}{.9}

\newcommand{\tool}{Vul-R2\xspace}
\newcommand{\moduleA}{DARL\xspace}
\newcommand{\moduleB}{CVRT\xspace}

\definecolor{lightgreen}{RGB}{0,150,0}
\newcommand{\scimp}[2]{%
  \ensuremath{#1\mathrlap{^{\textcolor{lightgreen}{\scalebox{0.8}{#2}}}}}%
}

\newcommand{\scdec}[2]{%
  \ensuremath{#1\mathrlap{^{\textcolor{gray!50}{\scalebox{0.8}{#2}}}}}%
}

\newcommand{\scdown}[2]{%
  \ensuremath{#1\mathrlap{^{\textcolor{red}{\scalebox{0.8}{#2}}}}}%
}

\newcommand{\http}{\url{https://github.com/Xin-Cheng-Wen/Vul-R2}}

\usepackage{tikz}

\def\BibTeX{{\rm B\kern-.05em{\sc i\kern-.025em b}\kern-.08em
    T\kern-.1667em\lower.7ex\hbox{E}\kern-.125emX}}
\begin{document}

\title{\raisebox{-0.1in}{\includegraphics[width=0.4in]{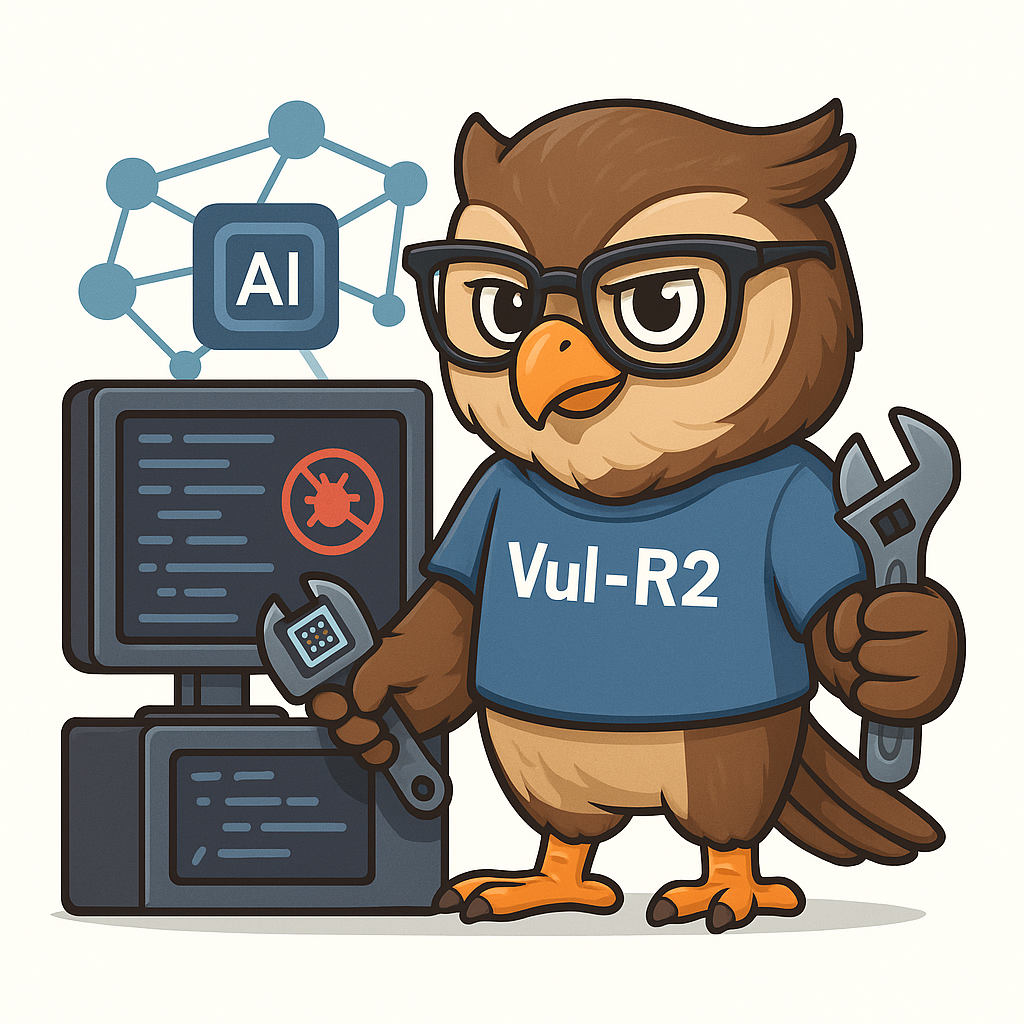}} Vul-R2: A Reasoning LLM for Automated Vulnerability Repair}



\author{\IEEEauthorblockN{Xin-Cheng Wen$^{1\dag}$, Zirui Lin$^{2}$, Yijun Yang$^{1}$, Cuiyun Gao$^{3\ast}$, Deheng Ye$^{1\ddag}$}

\IEEEauthorblockA{$^1$ Tencent Inc., Shenzhen, China}

\IEEEauthorblockA{$^2$ Department of Computer Science, City University of Hong Kong, China}

\IEEEauthorblockA{$^3$ Department of Computer Science and Engineering, The Chinese University of Hong Kong, China}

\IEEEauthorblockA{xiamenwxc@foxmail.com, ryanlzr2003@gmail.com, yijun.steven.yang@gmail.com, \\ cuiyungao@outlook.com,
dericye@tencent.com
}

\thanks{$^{\dag}$ Work done at Tencent Inc.}
\thanks{$^{\ast}$ Corresponding author.}
\thanks{$^{\ddag}$ Project leader.}
}

\maketitle

\begin{abstract}
The exponential increase in software vulnerabilities has created an urgent need for automatic vulnerability repair (AVR) solutions. Recent research has formulated AVR as a sequence generation problem and has leveraged large language models (LLMs) to address this problem.
Typically, these approaches prompt or fine-tune LLMs to generate repairs for vulnerabilities directly. Although these methods show state-of-the-art performance, they face the following challenges:
(1) Lack of high-quality, vulnerability-related reasoning data. Current approaches primarily rely on foundation models that mainly encode general programming knowledge. Without vulnerability-related reasoning data, they tend to fail to capture the diverse vulnerability repair patterns.
(2) Hard to verify the intermediate vulnerability repair process during LLM training.
Existing reinforcement learning methods often leverage intermediate execution feedback from the environment (e.g., sandbox-based execution results) to guide reinforcement learning training. In contrast, the vulnerability repair process generally lacks such intermediate, verifiable feedback, which poses additional challenges for model training. 

To address these challenges, we propose to model the vulnerability repair task from a reasoning perspective and train
a reasoning LLM termed \textit{Vulnerability Reasoner and Repair} (\tool)
which consists of two key modules:
(1) a domain-aware reasoning learning module, which comprises a reasoning answer construction component, a reasoning data filtering process, and a supervised fine-tuning process for learning vulnerability-related reasoning knowledge; and
(2) a curriculum-based verifiable rewarded training module,
which comprises dynamically reinforcement learning with verifiable rewards paradigms based on multiple-choice question answering in an easy stage and character-level matching in a hard stage. 
We evaluate \tool on the real-world C/C++ dataset PrimeVul to demonstrate its effectiveness in vulnerability repair. Specifically,  \tool outperforms the best baseline by 11.27\% for exact match (EM) and successfully repairs 49 additional vulnerabilities.
Furthermore, we demonstrate the effectiveness of the proposed paradigm, fine-tuning \tool on PrimeVul leads to improved EM performance of 8.78\% on a human curated dataset SVEN, even without additional training.

\end{abstract}

\begin{IEEEkeywords}
Vulnerability Repair, Large Language Model, Reinforcement Learning
\end{IEEEkeywords}

\section{Introduction}\label{sec:intro} 
The explosive complexity and scale of contemporary software systems have led to a significant increase in software vulnerabilities~\cite{DBLP:conf/kbse/WenWGWLG23}. Recent advancements in vulnerability detection~\cite{DBLP:journals/tse/WenGLWLL24} have further facilitated the identification and reporting of these vulnerabilities. According to the 2024 Synopsys report~\cite{Statista1}, 84\% of codebases contain at least one open-source vulnerability. Unaddressed vulnerabilities pose substantial risks, including financial loss, data breaches, and systemic failures~\cite{DBLP:conf/icse/CaoSBWLT22/mvd, reveal, devign, DBLP:conf/kbse/WangLPGCWGL23}. For instance, Cybersecurity Ventures projects that the global cost of cyberattacks will reach \$9.5 trillion in 2024, with ransomware, phishing, and data breaches constituting the primary drivers of this increase~\cite{news}. Despite the critical importance of timely vulnerability remediation, the repair process remains highly labor-intensive~\cite{DBLP:journals/corr/abs-2301-05456/dataquality}. Security experts must conduct root cause analyses, validate proposed fixes, and ensure compatibility across the codebase, often requiring hundreds of developer hours to address a single critical flaw.
Recent years have witnessed the emergence of numerous automated vulnerability repair (AVR) methods, which aim to automate the vulnerability remediation process and reduce the manual effort required from developers.

The earliest proposed  AVR techniques are rule-based methods~\cite{DBLP:conf/icse/GaoXMZYZXM15/repair1, DBLP:conf/icse/HongLLO20/repair2, DBLP:conf/sigsoft/LeeHO18/repair3, DBLP:journals/tdsc/LiLWMW24}. It predominantly employs static~\cite{CLANG, INFER} or dynamic analysis~\cite{fuzz} grounded in predefined rule sets. Although effective within their prescribed domains, the performance of these approaches is constrained by a limited coverage of vulnerability types, hindering their generalization to the diverse and continuously evolving landscape of software vulnerabilities in practice. 

\begin{figure}
    \centering
    \includegraphics[width=0.45 \textwidth]{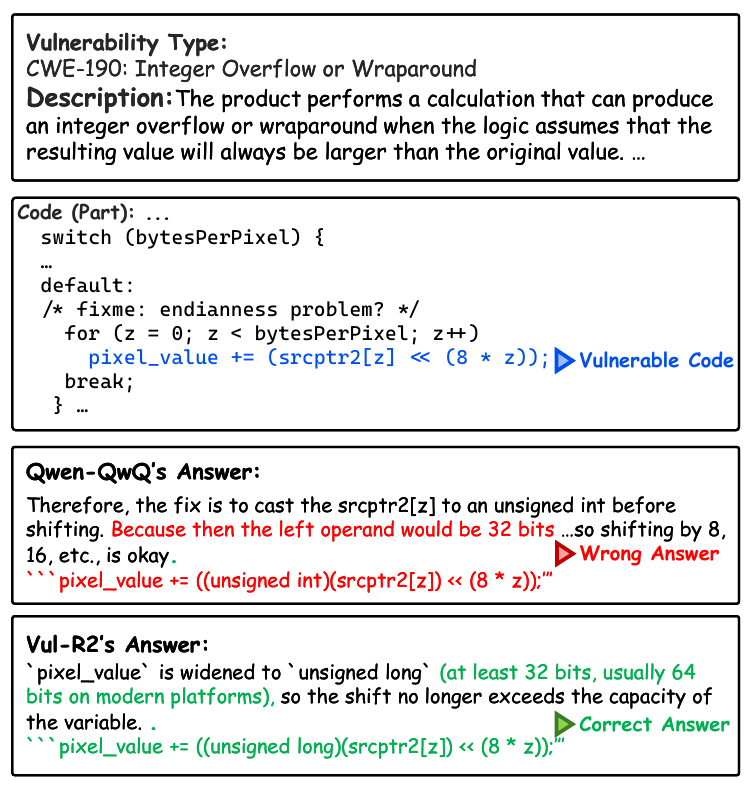}
    \caption{Illustration
    of the vulnerability ``Integer Overflow or Wraparound~\cite{CWE190}'' repaired by an open-source reasoning LLM (i.e., QwQ-32B~\cite{QwQ-32B}) and our Vul-R2. More detailed reasoning traces and the case of Vul-R2 can be found in Fig.~\ref{fig:detailed_case}. We adopt QwQ-32B in the case since the reasoning process of privileged LLMs, such as OpenAI-o3~\cite{gpto3}, is inaccessible due to their usage policies.\looseness-1}
    \label{fig:motivation}
    \vspace{-0.4cm}
\end{figure}

Inspired by the remarkable achievements of deep learning in the natural language processing (NLP) field~\cite{yang2024embodied, wei2025gtr, DBLP:conf/acl/WenYGXY25, DBLP:conf/issta/WenGGXL24, DBLP:journals/pacmpl/LiWWS25,gao2023makes}, code pre-trained models (CodePTMs) (i.e., CodeT5~\cite{DBLP:conf/emnlp/0034WJH21/CodeT5}) and large language models (LLMs) (i.e., ChatGPT~\cite{ChatGPT}) are widely adopted for AVR. CodePTM-based methods~\cite{DBLP:conf/sigsoft/FuTLNP22/vulrepair, DBLP:conf/icse/ZhouKXH024/vulmaster, DBLP:conf/kbse/WenWGWLG23} take the vulnerable code as input and generate the corresponding repaired code as output to address software vulnerabilities. They can naturally avoid the low coverage problem of rule-based approaches, as CodePTMs are rule-agnostic and make predictions based on probability~\cite{DBLP:conf/kbse/PengWWGL23}. 
Despite the effectiveness, CodePTM-based AVR approaches perform badly on rare types and are overly concerned with semantic information.
The recent emergence and rapid advancement of LLMs have substantially elevated the performance of automated code analysis and repair systems~\cite{DBLP:journals/corr/abs-2407-01489}, rendering them highly effective for AVR. LLM-based methods~\cite{DBLP:journals/corr/abs-2404-02525} have achieved improvements in repair accuracy and coverage. 
These approaches benefit from pre-training on large-scale code datasets~\cite{DBLP:journals/corr/abs-2401-16185/LLM4Vuln},  
pushing a rapid paradigm shift in the realm of vulnerability repair.
Despite the impressive success, they still face challenges in learning
the vulnerability repair patterns:

\textbf{(1) Lack of high-quality, vulnerability-related reasoning data.}
While recent efforts have sought to enhance LLMs through techniques such as retrieving similar code snippets via chain-of-thought (CoT)~\cite{DBLP:journals/corr/abs-2210-01240/cot} prompting or supervised fine-tuning (SFT)~\cite{DBLP:conf/acl/DongYLLXL00ZZ24/sft}, these approaches primarily rely on foundation models that mainly encode general programming knowledge. 
Due to the complex vulnerability trigger patterns,
the absence of explicit reasoning data tends to limit the model’s ability to
capture the diverse vulnerability repair strategies.
For example, as shown in Fig.~\ref{fig:motivation}, 
QwQ fails to capture the underlying root cause of vulnerability.
Specifically, since \texttt{pixel\_value} is a 64-bit variable, the data type of the shift operand should also be 64-bit to prevent integer overflow. However, QwQ incorrectly treated it as 32-bit during its reasoning process.
\textbf{(2) Hard to verify the intermediate vulnerability repair process during LLM training.} 
Effective vulnerability repair necessitates a multi-step planning process about value ranges and temporal relationships among symbolic variables~\cite{DBLP:journals/corr/abs-2403-17218}.
However, existing reinforcement learning-based approaches often leverage intermediate execution feedback from the environment (e.g., sandbox-based execution results) to guide training. The vulnerability repair process generally lacks such intermediate, verifiable feedback, which poses challenges for model learning.
As illustrated in Fig.~\ref{fig:motivation}, with the incorrect reasoning process, QWQ~\cite{QwQ-32B} produces a wrong result, which could be alleviated by intermediate verification.

\textbf{Our work.} To address the above challenges, we propose modeling the vulnerability repair task from a reasoning perspective, rather than relying on traditional prompt-based or SFT paradigms. 
Specifically, we formulate vulnerability repair as an exploratory-feedback problem, where the objective is to employ step-by-step reasoning to identify correct fixes from a large set of candidate solutions generated by LLMs, with the process being guided by verifiable feedback.
It closely mirrors the way humans learn and reason through interactive feedback.

Specifically, we propose a reasoning LLM for vulnerability repair named \textit{\tool}, which consists of two key components:
(1) A domain-aware reasoning learning module, 
which comprises a reasoning answer construction step for generating vulnerability-related reasoning data, a data filtering process to mitigate the impact of low-quality data, and an SFT process to bootstrap the model’s understanding of vulnerability-related concepts.
(2) A curriculum-based verifiable rewarded training module, enabling the model to progressively learn reasoning capabilities from an easy-to-hard stage.
In the easy stage, we design multiple-choice questions by dynamically constructing verifiable reward signals. In the hard stage, we utilize character-level matching to further enhance the model's repair capabilities through reinforcement learning with verifiable rewards (RLVR). 

To evaluate the effectiveness of \tool, we compare it with seven existing vulnerability repair baselines on the two high-quality C/C++ benchmark datasets: PrimeVul~\cite{DBLP:journals/corr/abs-2403-18624/primevul} and SVEN~\cite{DBLP:conf/ccs/HeV23/SVEN}. 
Experimental results show that \tool outperforms the best baseline by 11.27\% for exact match (EM) and successfully repairs 49 additional vulnerabilities in PrimeVul.
Furthermore, 
we validate the generalization capability of \tool across another dataset, 
training a model on PrimeVul can help improve performance on SVEN.


\textbf{Contributions.} The major contributions of this paper are summarized as follows:
\begin{enumerate}
\item 
To the best of our knowledge, this is the first work exploring
reasoning LLM for the vulnerability repair.

\item We propose \tool, a reasoning LLM for vulnerability repair. \tool effectively enables reasoning about complex vulnerability patterns through the domain-aware reasoning and curriculum-based verifiable rewarded training.

\item We extend the PrimeVul dataset with CoT reasoning answers for vulnerability repair and conduct an extensive evaluation.
The results demonstrate the effectiveness of \tool compared with baseline AVR methods. 

\end{enumerate}

\begin{figure*}[t]
	\centering
	\includegraphics[width=0.85\textwidth]{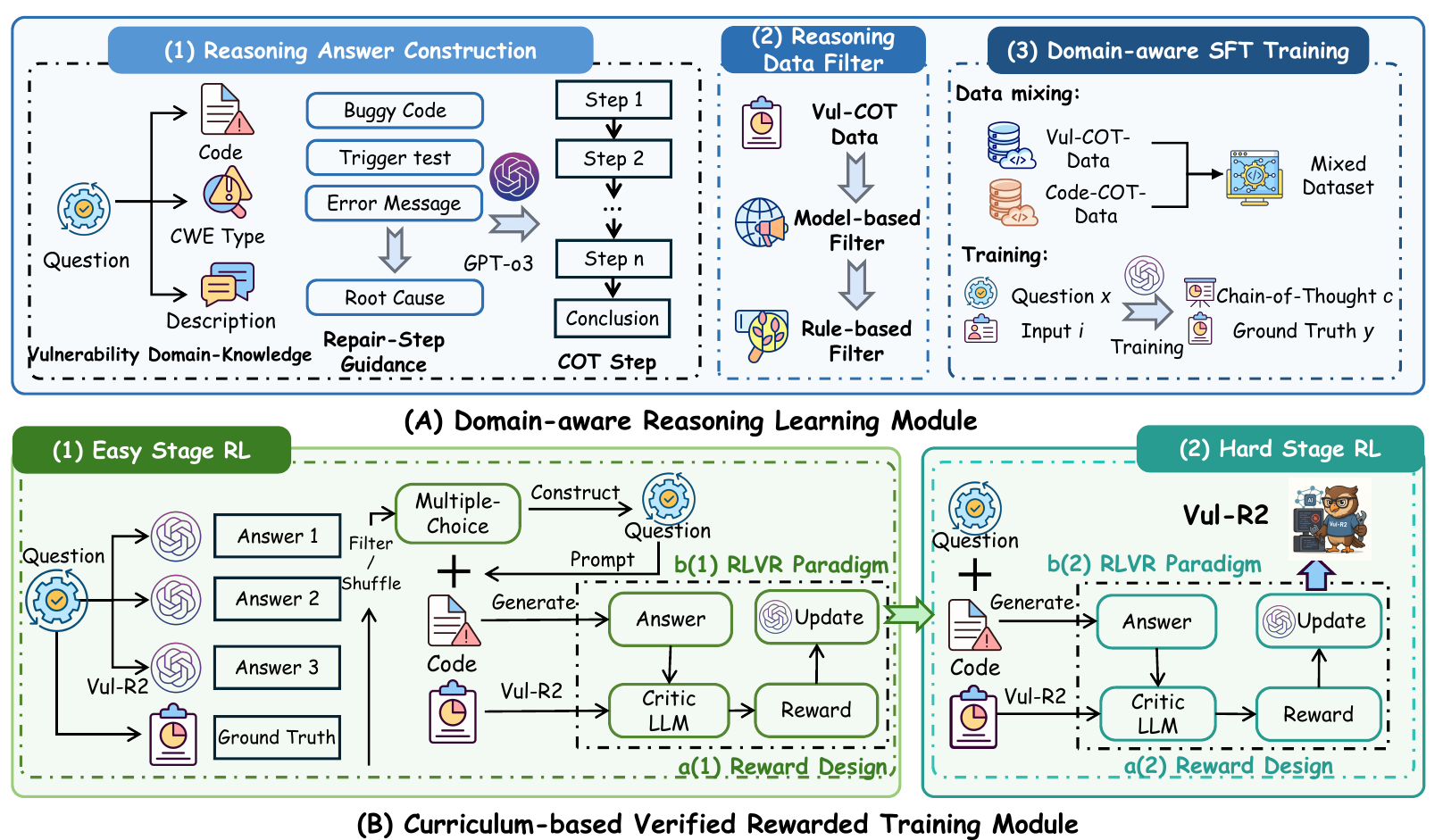}
    \caption{The overview of \tool.} 
    \vspace{-0.4cm}
\label{architecture}
\end{figure*}

\section{Proposed Framework}\label{sec:med}
\subsection{Overview}

Fig.~\ref{architecture} presents an overview of the proposed \tool. The primary objective of \tool is to employ a reasoning process, enabling the model to first acquire fundamental reasoning skills. Guided by reinforcement learning with verifiable feedback, the model then iteratively selects the optimal reasoning steps to accurately repair vulnerabilities, progressing from easy to hard stages.

\textbf{First,} the process begins with the domain-aware reasoning learning module, which serves as a cold-start phase to familiarize the model with the reasoning process. This module comprises three steps: the reasoning answer construction, which generates the vulnerability-related reasoning data;  the data filtering, which mitigates the impact of low-quality data; and an SFT training objective, which integrates domain-specific knowledge into the model.
\textbf{Subsequently}, \tool employs a curriculum-based verifiable rewarded training module. It consists of two stages, starting with multiple-choice questions in the easy stage and then advancing to complex vulnerability repair tasks in the hard stage. 

\subsection{Domain-Aware Reasoning Learning (\moduleA)}
We propose the domain-aware reasoning learning module
for facilitating the initial learning of vulnerability-specific reasoning knowledge. 
As shown in Fig.~\ref{architecture} (A),  it mainly contains three steps, including (a) reasoning answer construction,  (b) reasoning data filter,  and (c) domain-aware SFT training, with details as below.
\begin{figure}
    \centering
    \includegraphics[width=0.47 \textwidth]{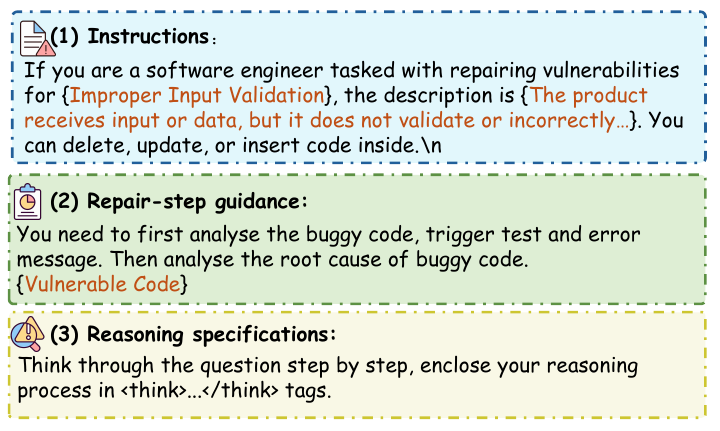}
    \caption{The illustration of the prompt in the RAC. Contents in ``\{ \}'' will be substituted by the corresponding data.}
    \label{fig:prompt}
    \vspace{-0.4cm}
\end{figure}
\paragraph{Reasoning Answer Construction (RAC)}
This component aims at generating the vulnerability-related reasoning data.
In real-world scenarios, developers' reasoning processes for identifying the root causes of vulnerabilities involve more than simply concluding directly.
To this end, we propose a reasoning prompt to generate the solution steps \(c\) for vulnerability repair.
Specifically, we construct the reasoning prompt as illustrated in Fig.\ref{fig:prompt} (1)-(3). The prompt comprises three components: (1) instructions incorporating vulnerability domain knowledge, (2) repair-step guidance, and (3) reasoning specifications coupled with inverse verification.

\textbf{(1) Instructions with domain knowledge:} 
It clarifies the task requirements and outlines the steps to be executed by the LLMs. We provide explicit reasoning instructions, encouraging the LLM to first internally deliberate on the reasoning process before presenting the final answer to the user. Additionally, we incorporate common weakness enumeration (CWE)~\cite{CWE} vulnerability information and detailed vulnerability descriptions.

\textbf{(2) Repair-step guidance:} We summarize three key steps to reason the root cause of vulnerability in the target sample: analyzing the buggy code, performing trigger tests, and examining error messages. 
Specifically, the model initially performs a thorough analysis of the annotated vulnerable code; subsequently, it applies trigger tests, such as boundary condition assessments; 
finally, it corroborates whether the observed outcomes correspond with the reported error messages.

\textbf{(3) Reasoning specifications with inverse verification:} It aims to standardize the output format. We provide reference answers to enable the model to perform a secondary verification of its own responses. LLMs are instructed to follow the given format step-by-step and produce results accordingly. Based on the reference answers, the model may overturn its previous conclusions and generate a more precise reasoning process. The verification step helps further mitigate hallucinations and improve the reliability of the model's outputs.

\paragraph{Reasoning Data Filter}
To ensure the quality of the constructed reasoning
data, we introduce the data filter process to prevent misleading the model during training. As shown in Fig.~\ref{architecture}, the data filter process mainly includes (a) model-based filtering and (b) rule-based filtering steps.

\textbf{Model-based filtering.} Inspired by prior work~\cite{DBLP:conf/iclr/Tong0ZC25}, the model-based filtering treats an LLM as a judge. In this process, the LLM functions as a binary classifier, responding with ``yes'' or ``no''. We retained those samples for which the model’s judgment aligned with the vulnerability type and description.

\textbf{Rule-based filtering.} Subsequently, rule-based filtering is applied to assess the format and content of the responses. The main rules are as follows: (1) We perform a comparison between the responses and the ground truth to ensure answer correctness. (2) Samples lacking intermediate reasoning steps are filtered out by using regular expressions, as they do not reflect the reasoning process. (3) We enforce a strict response format, where tags such as ``\texttt{<answer>...\texttt{</answer>}}'' denote the final answer and ``\texttt{<think>...\texttt{</think>}}'' indicate the reasoning process; Samples not conforming to this format are discarded. 
The filtered dataset is denoted as 
\(\mathcal{D_\text{vul}}\).
The detailed statistical results are listed in Table s1 on GitHub.

\paragraph{Domain-aware SFT Training}
Due to the limited vulnerability data availability,
we leverage a dataset of challenging algorithmic optimization problems by CodeForce~\cite{DBLP:journals/corr/abs-2501-01257/codeforce} (denoted as 
\(\mathcal{D_\text{code}}\)) to enhance the model’s code reasoning capabilities. Unlike the single-task vulnerability repair setting, we construct a mixed dataset \(\mathcal{D_\text{mixed}} = \mathcal{D_\text{vul}} \cup \mathcal{D_\text{code}}\). Each input includes a question \(x\) and additional contextual information \(i\) of vulnerability details for AVR, whereas algorithmic optimization problems include test cases. 
We utilize the CoT reasoning answers generated in the previous RAC step,
which is denoted as the solution steps \(c\). The answer serves as the ground truth \(y\). The model is then trained to fit these reference responses, comprising both the intermediate steps \(c\) and the ground truth \(y\) by maximum likelihood estimation as below:
\begin{align}
\mathcal{L}_{\text{SFT}} \;=\;
-\;\mathbb{E}_{{(x,i,c,y)\sim\mathcal{D}_{\text{mixed}}}}
\log \pi_\theta\;\bigl(c, y \mid x, i).
\end{align}

\subsection{Curriculum-based Verifiable Rewarded Training (\moduleB)}
This module is designed to train LLMs to acquire vulnerability-related reasoning capabilities and
find the best generated answer by dynamically constructing a verifiable reward signal. It consists of three components: (a) reward design, (b) the RLVR paradigm, and (c) two-stage RLVR. Notably, the two stages are optimized through the designed reward and
RLVR paradigm in the first two components. The complete process is presented in Algorithm 1.
\paragraph{Reward Design}
\label{Sec:reward}
Reward functions play a primary role in RLVR, shaping the landscape of policy optimization. Prior work~\cite{DBLP:journals/corr/abs-2502-14768/logicrl} has demonstrated that appropriately adjusting task difficulty is crucial for promoting training. To avoid reward hacking~\cite{DBLP:journals/corr/abs-2209-13085/rewardhacking}, we design answer and format rewards.


\begin{figure}
    \centering
    \includegraphics[width=0.47 \textwidth]{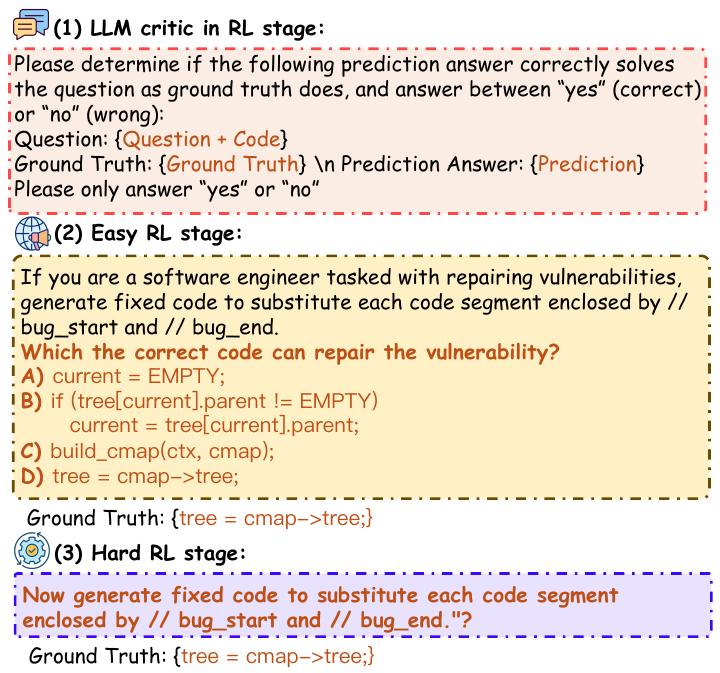}
    \caption{The illustration of the prompt in the \moduleB module.}
    \label{fig:prompt_rl}
    \vspace{-0.4cm}
\end{figure}
To evaluate the correctness of the generated answer \( o \), we employ an auxiliary LLM (i.e., Qwen2.5-14B-Instruct-1M~\cite{DBLP:journals/corr/abs-2501-15383/14B}) as a critic to assess whether the proposed answer effectively fixes the vulnerability. It mitigates the influence of irrelevant factors such as whitespace, comments, or unrelated variable names for evaluation. Specifically, the answer reward is calculated as follows:
\begin{equation}
\mathcal R(o)_{\text{acc}} = \begin{cases}
    -2, &\text{if the ``Critic'' is 0},\\ 
    \text{1}+\mathit{Sim}(\text{$o$}, \text{GT}), & \text{otherwise}.
\end{cases}
\label{eq:reward}
\end{equation}






\noindent where \(Sim\) reflects the closeness of the generated repair to the human-written ground truth patch (i.e., GT)~\cite{DBLP:journals/corr/abs-2502-18449/swerl}, which ranges from 0 to 1.  
The critic score (i.e., ``Critic'') of 1 is assigned if the critic LLM confirms that the answer correctly repairs the vulnerability. 
Formally, the critic LLM's prompt is structured as in Fig.~\ref{fig:prompt_rl} (1).

The format reward enforces output consistency and enhances the interpretability and post-processing of the model’s responses. It takes the value of 1 if the model's output strictly follows this predefined format. If the required tags are missing or incorrectly used, the format reward is set to -1. 

\definecolor{mydarkred}{RGB}{139, 0, 0}
\begin{algorithm}[!tpb]
    \SetAlgoLined
    \footnotesize
    \SetKwInOut{Input}{Input}
    \SetKwInOut{Output}{Output}
    \SetKwInOut{Initialize}{Initialize}
    \SetKwProg{Fn}{Function}{:}{}
    \Input{The original question $X$, the vulnerable code needs to be repaired, and the corresponding contextual information $I$, Answer $Y$, Model $\theta$, maximum generation number $J$.}
    
    \Fn{Curriculum-based Verifiable Rewarded Training}{
        
        // \textcolor{mydarkred}{Easy Stage RLVR}
        
        \For{all $x_n \in X$} 
        {
            Obtain  question $x_n$, context input $i_n$, and ground truth $y_n$
            
            j = 0
            
            \While{$x < J$} 
                 {Generate answer $x_j$ via $\theta$
                 
                 \If{Format correct}
                 {j += 1}
                 }                  
            Select answers samples $x_j$  and construct prompt $x_\text{easy}$ in Fig.~\ref{fig:prompt} (5)

            Optimization Process by Eq.~\ref{re+}, 5, 6, and 7 with $x_\text{easy}$, $i_n$, $y_n$

            Update Model $\theta$ 
        }
        // \textcolor{mydarkred}{Hard Stage RLVR}
        
        \For{all $x_n \in Q$} 
        {
            Obtain question $x_n$, context input $i_n$, and ground truth $y_n$
            
            Construct prompt $x_\text{hard}$ in Fig.~\ref{fig:prompt} (6)

            Optimization Process by Eq.~\ref{re+}, 5, 6, and 7 with $x_\text{hard}$, $i_n$, $y_n$
            
            Update Model $\theta$ 
        }
    }
    \Return {Model $\theta$}

\caption{The curriculum-based verifiable rewarded training process.}
\label{algo}
\vspace{-0.1cm}
\end{algorithm}

\paragraph{RLVR Paradigm}
\label{Sec:RLVR}
We use the same paradigm as inspired by REINFORCE++~\cite{DBLP:journals/corr/abs-2501-03262/re++} for two training phases. This can be viewed as a curriculum approach between easy and hard questions. 
To enhance training stability and efficiency between the two stages, we employ the KL divergence in the training objectives. 
Specifically, the objective function for sequence generation is formulated as follows:
\begin{align}
    \mathcal{J}_{\text{Re++}}(\theta) &= \mathbb{E}_{[q \sim P(Q), \{o_i\}_{i=1}^N \sim \pi_{\theta_{\text{old}}}(O|q)]} \notag \\
    &\frac{1}{N}\sum_{i=1}^N\frac{1}{\vert o_i\vert}\sum_{t=1}^{\vert o_i\vert}\Bigg\{\min \big[
    \frac{\pi_{\theta}^{i,t}}{\pi_{\theta_{\text{old}}}^{i,t}}\hat{A}_{i,t}, \notag \\
    & \textrm{clip}\left(\frac{\pi_{\theta}^{i,t}}{\pi_{\theta_{\text{old}}}^{i,t}}, 1-\epsilon, 1+\epsilon\right)\hat{A}_{i,t} 
    \big]
    \notag \\
    & - \beta\mathbb{D}_{\text{KL}}[\pi_{\theta} \| \pi_{\text{ref}}]\Bigg\}
    \label{re+}
\end{align}

\noindent where \( P(Q) \) represents the distribution of vulnerability repair questions used during training, with \( q \) denoting a sample drawn in the current training iteration. The old policy and the current policy of base model are represented by \( \pi_{\theta_{\text{old}}} \) and \( \pi_{\theta_{\text{new}}} \), respectively, where \( o \) corresponds to a complete response sampled from the respective policy. The reference policy, corresponding to the frozen base model parameters, is denoted by \( \pi_{\theta_{\text{ref}}} \). Additionally, \( N \) indicates the number of responses sampled per question in each iteration and $\epsilon$ is the clipping threshold for policy updates. \( \hat{A}_{i,t} \) denotes the advantage function, which is defined as: 
\begin{align}
\hat{A}_{i,t} =  \frac{r_i - \text{mean}(\{r_1, r_2, \cdots, r_N\})}{\text{std}(\{r_1, r_2, \cdots, r_N\})} \\
r_i = r(x, y) - \beta \cdot \sum_{i=t}^{T} \mathbb{D}_{\text{KL}}\left[\pi_{\theta} \| \pi_{\text{ref}}\right]
\end{align}
where \( \{r_1, r_2, \cdots, r_N \}\) denotes a group of rewards corresponding to the outputs within each group.
\(\mathbb{D}_{\text{KL}}\left[\pi_{\theta} \| \pi_{\text{ref}}\right]\) denotes the KL divergence \cite{DBLP:conf/nips/Ouyang0JAWMZASR22/kl}, which employs an unbiased estimator and is formulated as follows:
\begin{equation}
\mathbb{D}_{\text{KL}}\left[\pi_{\theta} \| \pi_{\text{ref}}\right] = \frac{\pi_{\text{ref}}\left(o_{i, t} | q, o_{i,<t}\right)}{\pi_{\theta}\left(o_{i, t} | q, o_{i,<t}\right)} - \log \frac{\pi_{\text{ref}}\left(o_{i, t} | q, o_{i,<t}\right)}{\pi_{\theta}\left(o_{i, t} | q, o_{i,<t}\right)} - 1.
\end{equation}
where \( \frac{\pi_{\theta}(o_{i,t} | q)}{\pi_{\theta_{\text{ref}}}(o_{i,t} | q)} \) is the
policy ratio.  
It regularizes the policy update, ensuring that \( \pi_{\theta} \) does not deviate excessively from the reference model \( \pi_{\theta_{\text{ref}}} \).

\paragraph{Two Stage RLVR}
Directly applying the existing RLVR paradigm to software vulnerability repair is generally ineffective. 
This is primarily due to the 
the vulnerability repair process does not produce verifiable intermediate feedback, which hinders its ability to generate accurate vulnerability fixes. 
Specifically, we propose a two-stage RLVR process comprising an easy stage and a hard stage, as detailed in Algorithm~\ref{algo}.
\textbf{(1) Easy Stage:}
We reformulate the generation task as a multiple-choice problem. Specifically, to initialize the easy stage RL (Lines 3-15 in Algorithm~\ref{algo}). Each prompt samples up to \(J\) from the
current dataset \(\mathcal{D}_\text{vul}\) as candidate vulnerability repair answers. These approximate incorrect answers \(x_j\), together with the ground truth, constitute options A, B, C, and D to construct multiple-choice prompt \(x_\text{easy}\) in Fig.~\ref{fig:prompt_rl} (2). 
The model \(\theta\) is then instructed to output the single-letter choice, along with the corresponding ground truth \(y\) enclosed within ``\texttt{<answer>...\texttt{</answer>}}'' tags. 
This approach encourages the model to explore solution paths that are similar to the correct answer, thereby reducing the likelihood of generating entirely invalid responses. 
Importantly, the model is not limited to generating only the single-letter choice; it also produces the complete repair code, which is valuable for the subsequent training stage.
\textbf{(2) Hard Stage:} In this stage, we continue training the model that was previously trained in the easy stage, without providing multiple-choice prompts, allowing the model to perform the vulnerability repair task in a more open-ended manner. During this phase, we encourage the model to engage in unrestricted exploration, which facilitates the acquisition of broader knowledge. The detailed prompt is shown in Fig.~\ref{fig:prompt_rl} (3). The prompts used in this stage are consistent with those employed during the inference phase, comprising the question \(x_{\text{hard}}\), input \(i\), and ground truth \(y\) components. The whole process is elaborated in our Algorithm~\ref{algo}.
The detailed data statistics are provided in Table s2 on the GitHub repository.

\section{Experimental setup}\label{sec:setup}
\subsection{Research Questions}

In this section, we evaluate the effectiveness of \tool by comparing it with the state-of-the-art baselines and focus on the following six research questions (RQs):

\begin{enumerate}[label=\bfseries RQ\arabic*:,leftmargin=.5in]
    \item How effective is \tool compared with existing vulnerability repair approaches?
     \item  How effective is \tool in repairing vulnerabilities across different vulnerability types?
      \item How do RLVR techniques contribute to the performance of \tool?
       \item  What is the impact of incorporating different reasoning data during the SFT phase?
    \item What is the influence of different components of \tool on the performance for repairing vulnerabilities? 
    \item What is the influence of hyperparameters on the performance of \tool?
\end{enumerate}

\subsection{Datasets}
In this paper, we focus on C/C++ programs due to their widespread adoption in real-world software development and the prevalence of well-known vulnerabilities that have been accurately labeled by security researchers. Specifically, we utilize two widely-used and high-quality datasets: PrimeVul~\cite{DBLP:journals/corr/abs-2403-18624/primevul} and SVEN~\cite{DBLP:conf/ccs/HeV23/SVEN}.

\paragraph{PrimeVul} PrimeVul is a large-scale real-world C/C++ dataset. 
It aggregates vulnerability-related commits from four established datasets, BigVul~\cite{BigVul}, CVEFixes~\cite{CVEFixes}, CrossVul~\cite{CrossVul}, and DiverseVul~\cite{DiverseVul}, covering more than 140 CWE categories. 
It applies a stringent deduplication process and adopts a temporally-aware data splitting strategy, resulting in training, validation, and test sets containing 3,789, 480, and 435 samples, respectively.


\paragraph{SVEN} SVEN dataset manually vets vulnerabilities
from multiple repositories of C/C++ code. It consists of 384 C/C++ vulnerabilities, covering 9 CWEs. By manual inspection, the accuracy of SVEN reaches 94.0\%.  We use SVEN as the test set without further training.

\subsection{Baselines}
To provide a comprehensive evaluation, we experiment on two types of methods, with details as below.
\textit{(a) CodePTM-based methods:} 
We choose two recent CodePTM-based works, VulRepair~\cite{DBLP:conf/sigsoft/FuTLNP22/vulrepair} and VulMaster~\cite{DBLP:conf/icse/ZhouKXH024/vulmaster}, as baselines. 
VulRepair fine-tunes CodeT5 with BPE tokenization~\cite{BPE} to improve automated patch generation accuracy.
VulMaster incorporates CWE knowledge and code structure into CodeT5 for AVR.
\textit{(b) LLM-based methods:} 
We select three large open-source LLMs: Llama3-70B-Instruct~\cite{DBLP:journals/corr/abs-2407-21783/llama3.1}, Qwen2.5-14B-Instruct-1M~\cite{DBLP:journals/corr/abs-2501-15383/14B}, and Qwen2.5-Coder-32B-Instruct~\cite{DBLP:journals/corr/abs-2409-12186/qwen2.5}, due to their strong performance in code generation tasks.
In addition, we incorporate the closed-source LLM OpenAI-o3~\cite{gpto3} for vulnerability repair, given its demonstrated effectiveness in handling general tasks.
\subsection{Metrics}
Following the previous methods~\cite{DBLP:conf/icse/ZhouKXH024/vulmaster, DBLP:conf/sigsoft/FuTLNP22/vulrepair}, we employ the EM, Success, and CodeBLEU metrics as evaluation metrics:
\paragraph{Exact Match and Success} Exact match (EM) is the percentage of generated fixes that match the token sequence of the ground truth, which means this part of the fixes is correct. 
It is formulated as follows:
$EM=\frac{\text{Correct Fixes}}{\text{ Vulnerabilities}}$.
Success metric equals the number of vulnerabilities that are successfully fixed.
\paragraph{CodeBLEU} CodeBLEU\cite{CodeBLEU} is used to evaluate the similarity between the generated code and the ground truth. It's a variant of BLEU\cite{BLEU} specialized for code tasks, which also considers the structural similarity of source code. 
\subsection{Implementation Details}

During the inference phase, we set the maximum beam size to 10 for most baseline implementations, except for OpenAI-o3\cite{gpto3}, where the beam size is set to 5. This choice is reasonable, as our beam size is comparable to or smaller than those used in prior work~\cite{huang2024template,gao2024search}.

In the \moduleA module, we employ OpenAI-o3~\cite{gpto3} as the data generation model for the RAC phase. For the SFT phase, we train Qwen-14B-Instruct-1M\cite{DBLP:journals/corr/abs-2501-15383/14B} with a batch size of 16, using a constant learning rate of 1e$^{-4}$ and the AdamW optimizer. The maximum prompt and response lengths are both set to 4,096 tokens. Additional training settings include gradient accumulation steps of 16, a cosine learning rate scheduler, three training epochs, and a warmup ratio of 0.1.

For the \moduleB phase, the training batch size is set to 16, with a maximum prompt length of 1,024 and a maximum response length of 4,096. The actor’s learning rate is set to 2e$^{-6}$.  The rollout temperature is set to 1.0. We generate 8 and 16 rollouts per question for RL reward computation in the easy and hard stages, respectively. 

During the \moduleA stage, we utilized 8 NVIDIA H20 GPUs. For the \moduleB stage, we employed 16 H20 GPUs for training. The detailed training set is available at GitHub.
\section{Experimental Results}\label{sec:result}

\begin{table*}[t]
\centering
\setlength{\tabcolsep}{2mm}
\renewcommand{\arraystretch}{1.1}
\caption{Experimental results of \tool and the vulnerability repair baselines on the PrimeVul and SVEN datasets. 
Texts in bold represent the best performance of the best methods in each metric.}
\resizebox{0.92\textwidth}{!}{
\begin{tabular}{c|l|c|c|cccccc}
\toprule
\rowcolor[HTML]{DEDEDE}
&  & \multicolumn{2}{c}{\textbf{Dataset}} & \multicolumn{3}{c}{\textbf{PrimeVul}~\cite{DBLP:journals/corr/abs-2403-18624/primevul}}  & \multicolumn{3}{c}{\textbf{SVEN}~\cite{DBLP:conf/ccs/HeV23/SVEN}}      \\
\rowcolor[HTML]{DEDEDE}
\multirow{-2}{*}{\textbf{Type}} &   \multirow{-2}{*}{\textbf{Method}}                        & \multicolumn{1}{l|}{\textbf{Paradigm}}   &  \multicolumn{1}{c|}{\textbf{Sample Size}}& \textbf{Success$\uparrow$} & \textbf{Exact Match $\uparrow$}   & \textbf{CodeBLEU$\uparrow$}  & \textbf{Success$\uparrow$} & \textbf{Exact Match$\uparrow$}   & \textbf{CodeBLEU$\uparrow$}  \\

\midrule
&{VulMaster}                     & SFT     & -              & 20                   & 4.59                 & 36.74                & 27      & 7.34  & 47.45    \\
               \multirow{-2}{*}{{CodePTM}} &          VulRepair                      & SFT        & 10             &                      8&                      1.84&                      32.86&         8&       2.17&          36.77\\
               \hline
&Qwen2.5-14B-Instruct                               & COT        & {10}       &  43& 9.89&  35.89&         42&       11.41&          41.73\\
& Qwen2.5-32B-Instruct& COT& 10& 51& 11.72& 34.79& 45& 12.23&41.45\\
&LLama3-70B-Instruct                             & COT        & 10             &                      24&                      5.52&                      34.47&         7&       3.95&          40.33\\
&OpenAI-o3                                             & COT        & 5              & 32                   & 7.36                 & 27.32                & 30      & 8.17  & 38.78    \\
\multirow{-5}{*}{LLM} &Qwen2.5-14B-Instruct& SFT        & 10             & 59                   & 13.56                & 39.20                & 112     & 30.35 & 56.68    \\
\bottomrule
\rowcolor[HTML]{ECECEC}
& \textbf{$\text{\tool}^{*}$} &  RLVR         & 10      & \scimp{\textbf{103}}{$\uparrow$\textbf{44}} & \scimp{\textbf{23.67}}{$\uparrow$\textbf{10.11\%}}  & \scimp{\textbf{42.95}}{$\uparrow$\textbf{3.75\%}} & \scimp{\textbf{142}}{$\uparrow$\textbf{30}} & \scimp{\textbf{38.59}}{$\uparrow$\textbf{8.24\%}} & \scimp{\textbf{60.07}}{$\uparrow$\textbf{3.39\%}} \\
\rowcolor[HTML]{DEDEDE}
\multirow{-2}{*}{\textbf{{Ours}}}  & \textbf{\tool} & SFT $\&$ RLVR         & 10      & \scimp{\textbf{108}}{$\uparrow$\textbf{49}} & \scimp{\textbf{24.83}}{$\uparrow$\textbf{11.27\%}}  & \scimp{\textbf{46.17}}{$\uparrow$\textbf{6.97\%}} & \scimp{\textbf{144}}{$\uparrow$\textbf{32}} & \scimp{\textbf{39.13}}{$\uparrow$\textbf{8.78\%}} & \scimp{\textbf{61.40}}{$\uparrow$\textbf{4.72\%}} \\
\bottomrule
\end{tabular}
}
\label{RQ1}
\vspace{-0.4cm}
\end{table*}

\subsection{RQ1: Comparison with SOTA}
To assess the effectiveness of \tool in improving the performance of vulnerability repair, we compare it
with seven baselines. Table~\ref{RQ1} presents the performance of \tool with baselines on PrimeVul and SVEN datasets.
\subsubsection{\tool vs CodePTMs}
The results summarized in Table~\ref{RQ1} demonstrate that \tool consistently outperforms all SFT-based CodePTMs methods across both evaluation datasets. Specifically, \tool achieves an average improvement of 28.00\% in EM and 15.33\% in CodeBLEU, respectively. 
Furthermore, on the SVEN dataset, where no additional training is conducted, \tool is able to correctly repair 144 vulnerabilities and achieves an EM of 39.13\%. In comparison, Vulmaster and VulRepair are only able to repair 27 and 8 vulnerabilities, respectively. These results indicate that the reasoning-based method adopted by \tool demonstrates stronger generalization capability on previously unseen datasets, which contributes to its superior performance.

\subsubsection{\tool vs LLMs}
As shown in Table~\ref{RQ1}, \tool consistently outperforms all LLM-based methods across all datasets and evaluation metrics, even when compared to generally robust models such as OpenAI-o3. Specifically, \tool achieves an EM score of up to 24.83\% and a CodeBLEU score of 46.17\% on the PrimeVul dataset. Moreover, on the SVEN dataset,  \tool exhibits a higher EM of 39.13\% and CodeBLEU of 61.40\%, further demonstrating its superior performance.
There are two primary factors that may contribute to this performance gap. First, the reasoning-based paradigm adopted by \tool generates a multi-step, verifiable, and high-quality repair process, which is ignored by
previous paradigms such as CoT prompting or SFT. Second, the LLM-based approaches may lack sufficient domain-specific knowledge required for effective vulnerability repair, making it particularly challenging to detect and address complex vulnerability patterns,
even when advanced reasoning capabilities are present in models such as OpenAI-o3.

\begin{tcolorbox}[width=\linewidth,boxrule=0pt,top=1pt, bottom=1pt, left=1pt,right=1pt, colback=gray!20,colframe=gray!20]
\textbf{Answer to RQ1:} 
\tool achieves the best overall performance across all evaluated metrics, with improvements of 8.78$\sim$36.96\% in EM and the identification of 32$\sim$137 additional vulnerabilities on the PrimeVul and SVEN datasets.
\end{tcolorbox}

\subsection{RQ2: Comparison with Vulnerability Types}
\begin{table}[t]
\centering
\setlength{\tabcolsep}{2mm}
\renewcommand{\arraystretch}{1.1}
\caption{Number of exact matches for the fine-grained CWE vulnerability types.} 
\resizebox{0.49\textwidth}{!}{
\begin{tabular}{l|rrrcc}
\toprule
\rowcolor[HTML]{DEDEDE}
\textbf{CWE-Type Category}                                    & \multicolumn{1}{r}{\textbf{VulMaster}} & \multicolumn{1}{r}{\textbf{OpenAI-o3}} & \multicolumn{1}{r}{\textbf{\tool}} & \textbf{EM $\uparrow$} & \space\space \\
\midrule
Resource Management Error                            & 9/229                                  & 15/229                 & 59/229                  & \scimp{25.76}{+19.21\%}  \\
\begin{tabular}[l]{@{}c@{}}Improper Check or Handling of \\ Exceptional Conditions\end{tabular} & 0/52                                   & 1/52                   & 9/52                   & \scimp{17.31}{+15.38\%}    \\
NULL Pointer Dereference                             & 2/39                                   & 6/39                   & 11/39                   & \scimp{28.21}{+12.82\%}    \\
Numeric Error                                        & 4/35                                   & 4/35                   & 10/35                   &  \scimp{28.57}{+17.14\%}   \\
Insufficient Control Flow Management                 & 3/26                                   & 0/26                   & 8/26                    & \scimp{30.77}{+19.23\%}    \\
Improper Input Validation                            & 0/24                                   & 1/24                   & 4/24                    & \scimp{16.67}{+12.50\%}    \\

Improper Access Control                              & 1/19                                   & 4/19                   & 6/19                    & \scimp{31.58}{+10.53\%}    \\

 Injection                                            & 0/4                                    & 0/4                    & 0/4                     &  \scdec{0.00}{+0\%}     \\
Others                                               & 1/11                                   & 1/11                   & 1/11                    & \scdec{9.09}{+0\%} 
      \\ 
\bottomrule
\end{tabular}
}
\label{RQ2}
\vspace{-0.4cm}
\end{table}
To evaluate the effectiveness of \tool in repairing different types of vulnerabilities, we select nine categories and 62 types of CWE across 143 projects. The proportion and types of vulnerabilities are presented in Table~\ref{RQ2}, and the selection criteria are based on CWE~\cite{CWE}.
It is important to note that each category may contain multiple specific CWE types.

Overall, we observe that \tool is effective across all the categories of vulnerabilities analyzed, achieving an average performance improvement of 18.04\% in EM. Specifically, we can observe the following findings:
(1) \tool demonstrates excellent performance in repairing vulnerabilities associated with Numeric Errors, correctly repairing 10 out of 35 vulnerabilities, resulting in an accuracy of 28.57\%. This could
be attributed to the RLVR paradigm, which excels at diverse reasoning skills under verifiable rewards, including arithmetic and logic~\cite{DBLP:journals/corr/abs-2402-03300/DeepSeekMath}.
(2) In addition, for the majority of vulnerability categories, \tool repairs more vulnerabilities than baselines. For example, in categories such as NULL Pointer Dereference, Improper Access Control, and Resource Management Error, \tool achieves accuracy rates of 28.21\%, 31.58\%, and 25.76\%, respectively.
(3) For vulnerability categories with limited training data, such as Injection vulnerabilities, \tool finds it challenging to achieve performance improvements. This is primarily because the same type of vulnerability can be triggered through various mechanisms, making it difficult to enhance performance when data is scarce.
\begin{tcolorbox}[width=\linewidth,boxrule=0pt,top=1pt, bottom=1pt, left=1pt,right=1pt, colback=gray!20,colframe=gray!20]
\textbf{Answer to RQ2:} 
Across most vulnerability categories, \tool successfully repairs more cases than other baseline methods. 
\tool particularly demonstrates strong performance in repairing vulnerabilities such as Numeric Errors.
\end{tcolorbox}

\subsection{RQ3: Analysis of RLVR}
To answer this RQ, we investigate the impact of the RLVR in \tool on AVR performance, and explore whether RLVR can trigger ``aha moment''~\cite{DBLP:journals/corr/abs-2501-12948/DeepSeekR1} similar to those observed in DeepSeek-R1. Due to space limitations, we present only a subset of the experimental results and training curves.

\begin{table}[t]
\centering
\setlength{\tabcolsep}{3mm}
\renewcommand{\arraystretch}{1.2}
\caption{Performance of \tool with respect to different RLVR-based methods.}
\resizebox{0.49\textwidth}{!}{
\begin{tabular}{l|cccccc}
\toprule

\rowcolor[HTML]{DEDEDE}
  & \multicolumn{3}{c}{\textbf{PrimeVul}~\cite{DBLP:journals/corr/abs-2403-18624/primevul}}  & \multicolumn{3}{c}{\textbf{SVEN}~\cite{DBLP:conf/ccs/HeV23/SVEN}}      \\
\rowcolor[HTML]{DEDEDE}
   \multirow{-2}{*}{\textbf{Method}}                        & \textbf{Success $\uparrow$} & \textbf{EM$\uparrow$}   & \textbf{CodeBLEU $\uparrow$}  & \textbf{Success $\uparrow$} & \textbf{EM$\uparrow$}   & \textbf{CodeBLEU $\uparrow$}  \\

\midrule
\rowcolor[HTML]{ECECEC}
\tool & \textbf{108} & \textbf{24.83} & \textbf{46.17} & \textbf{144} & \textbf{39.13} & \textbf{61.40} \\
\midrule

\textit{- w/ GRPO} & \scdown{98}{$\downarrow$-10} & \scdown{22.53}{$\downarrow$-2.30\%} & \scdown{43.79}{$\downarrow$-2.38\%} & \scdown{135}{$\downarrow$-9} & \scdown{36.68}{$\downarrow$-2.45\%} & \scdown{60.57}{$\downarrow$-0.83\%} \\
\textit{- w/o Critic LLM} & \scdown{101}{$\downarrow$-7} & \scdown{23.22}{$\downarrow$-1.61\%} & \scdown{45.00}{$\downarrow$-1.17\%} & \scdown{143}{$\downarrow$-1} & \scdown{38.86}{$\downarrow$-0.27\%} & \scdown{61.17}{$\downarrow$-0.23\%} \\
\textit{- w/ zero-RLVR} & \scdown{103}{$\downarrow$-5} & \scdown{23.67}{$\downarrow$-1.16\%} & \scdown{42.95}{$\downarrow$-3.22\%} & \scdown{142}{$\downarrow$-2} & \scdown{38.59}{$\downarrow$-0.54\%} & \scdown{60.07}{$\downarrow$-1.33\%} \\
 
\bottomrule
\end{tabular}
}
\label{RQ3}

\end{table}

\begin{table}[t]
\centering
\setlength{\tabcolsep}{2mm}
\renewcommand{\arraystretch}{1.2}
\caption{Performance of \tool with respect to different reasoning data domains.}
\resizebox{0.49\textwidth}{!}{
\begin{tabular}{l|cccccc}
\toprule
\rowcolor[HTML]{DEDEDE}
  & \multicolumn{3}{c}{\textbf{PrimeVul}~\cite{DBLP:journals/corr/abs-2403-18624/primevul}}  & \multicolumn{3}{c}{\textbf{SVEN}~\cite{DBLP:conf/ccs/HeV23/SVEN}}      \\
\rowcolor[HTML]{DEDEDE}
   \multirow{-2}{*}{\textbf{Data Type}}                        & \textbf{Success $\uparrow$} & \textbf{EM$\uparrow$}   & \textbf{CodeBLEU $\uparrow$}  & \textbf{Success $\uparrow$} & \textbf{EM$\uparrow$}   & \textbf{CodeBLEU $\uparrow$}  \\
\midrule
\rowcolor[HTML]{DCDCDC}
\textit{- w/o Reason}                                        & 59       & 13.56  & 39.20    & 112     & 30.35 & 56.68    \\
\midrule
\textit{- w/ Reason}        & \scimp{75}{$\uparrow$16}     & \scimp{17.24}{$\uparrow$3.68\%} & \scdec{39.20}{0.00\%} & \scdown{106}{$\downarrow$-6}   & \scdown{28.80}{$\downarrow$-1.55\%} & \scdown{54.41}{$\downarrow$-2.27\%} \\
\textit{- w/ Reason \& Math}        & \scimp{79}{$\uparrow$20}     & \scimp{18.16}{$\uparrow$4.60\%} & \scdown{36.49}{$\downarrow$-2.71\%} & \scdown{108}{$\downarrow$-4}   & \scdown{29.35}{$\downarrow$-1.00\%} & \scdown{47.55}{$\downarrow$-9.13\%} \\
\textit{- w/ Reason \& Code}        & \scimp{\textbf{97}}{$\uparrow$\textbf{38}} & \scimp{\textbf{22.30}}{$\uparrow$\textbf{8.74\%}}  & \scimp{\textbf{43.87}}{$\uparrow$\textbf{4.67\%}} & \scimp{\textbf{137}}{$\uparrow$\textbf{25}} & \scimp{\textbf{37.23}}{$\uparrow$\textbf{6.88\%}} & \scimp{\textbf{61.12}}{$\uparrow$\textbf{4.44\%}} \\
\bottomrule
\end{tabular}
}
\vspace{-0.4cm}
\label{RQ4_type}
\end{table}
\subsubsection{Effectiveness of RLVR} We design three variants:
(1) a variant that replaces the \moduleB component with GRPO, which is the RLVR method employed by DeepSeek-R1~\cite{DBLP:journals/corr/abs-2501-12948/DeepSeekR1}) (i.e., \textit{- w/ GRPO}).
(2) a variant of \tool in which the \moduleB operates without a critic LLM. Instead, the reward signal is computed solely based on rule-based correctness criteria (i.e., \textit{- w/o Critic LLM}).
(3) a variant that applies the RLVR paradigm directly, by passing the initial SFT phase (i.e., \textit{- w/ zero-RLVR}).

As shown in Table~\ref{RQ3}, the \moduleB component consistently outperforms GRPO across 2.3\% of EM and 2.38\% of CodeBLEU in PrimeVul. This observation is consistent with prior findings reported by Xie et al.~\cite{DBLP:journals/corr/abs-2502-14768/logicrl}.
Additionally, we find that the inclusion of a critic LLM has a substantial impact on performance. This may be attributed to the inherently diverse and non-deterministic nature of vulnerability repair, where multiple correct solutions often exist. A critic LLM is capable of assessing finer-grained aspects of generated patches, 
that rule-based metrics tend to overlook, thereby providing more discriminative reward signals during training.
Finally, we observe that even when used  RLVR paradigm independently, the RLVR paradigm yields competitive results, achieving an EM of 31.13\% and a CodeBLEU score of 51.51\%. These results collectively highlight the effectiveness of the RLVR paradigm in enhancing the reasoning capabilities of LLMs for AVR.

\begin{figure}
    \centering
    \includegraphics[width=0.50\textwidth]{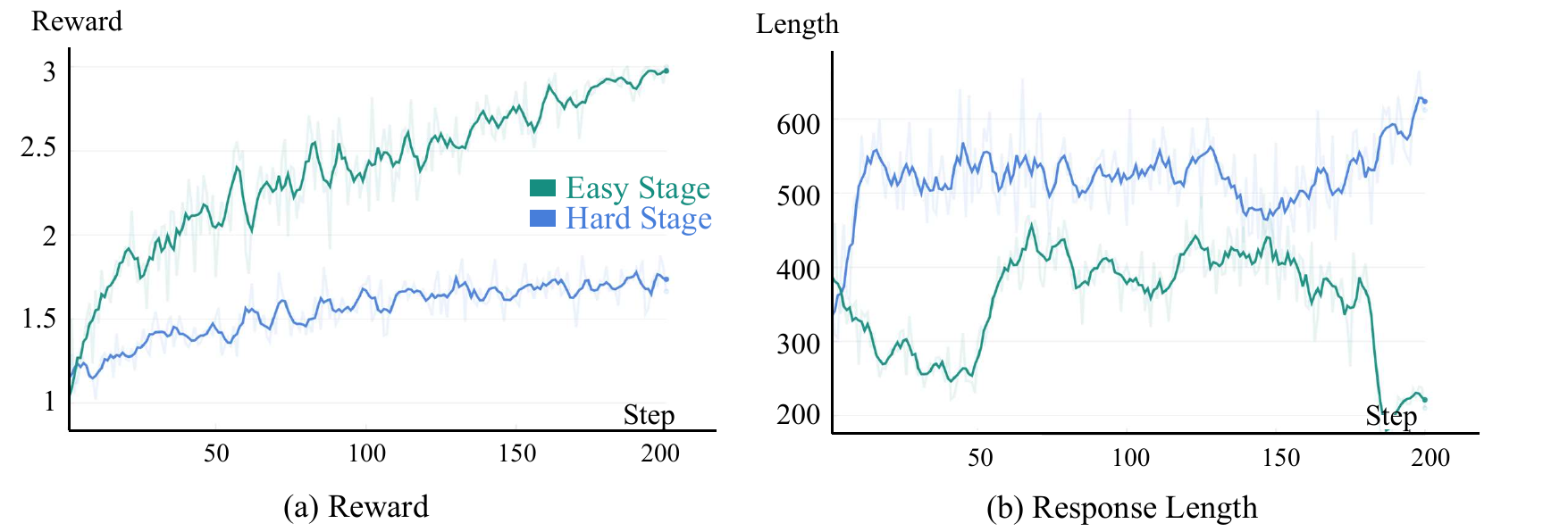}
    \caption{Reward and mean response length during RLVR training ($\text{\tool}^*$), illustrating how the model autonomously learns to allocate more thinking compute. 
    }
    \label{fig:aha}
    \vspace{-0.4cm}
\end{figure}

\subsubsection{Aha Moment in AVR}
We observe clear signs of ``aha moment'', as previously described by DeepSeekAI~\cite{DBLP:journals/corr/abs-2501-12948/DeepSeekR1}, wherein \tool demonstrates emergent reasoning abilities during the vulnerability repair process. To the best of our knowledge, this is the first empirical study to reveal such phenomena within the context of real-world, vulnerability-related tasks, extending the findings of DeepSeek-R1~\cite{DBLP:journals/corr/abs-2501-12948/DeepSeekR1}.

As illustrated in Fig.~\ref{fig:aha} (a), the easy stage learn the reward signal more effectively. Performance in both stages improves progressively during model training.
In Fig.~\ref{fig:aha} (b), the $\text{\tool}^*$ leads to increased average response lengths in the hard stage (i.e., blue lines), with the response length rising from 300 to 500. It suggests that the model allocates more ``thinking time'' to reflect upon its initial assumptions before arriving at a final repair. This reflective behavior emerges through interaction with the RLVR process and is not prompted by explicit instructions. We also observe a noticeable rise in the usage of introspective linguistic markers, such as ``verify'' and ``ensure'', which often indicate nuanced reasoning. 
Such patterns suggest the development of a more deliberate reasoning process. 

\begin{tcolorbox}[width=\linewidth,boxrule=0pt,top=1pt, bottom=1pt, left=1pt,right=1pt, colback=gray!20,colframe=gray!20]
\textbf{Answer to RQ3:} 
Our results provide the first empirical evidence that the RLVR technique improves the performance of reasoning capabilities in vulnerability-related tasks, as indicated by the emergence of an observable ``aha moment" within the vulnerability domain. 
\end{tcolorbox}

\subsection{RQ4: Analysis of Reasoning Data}
To systematically explore the contribution of the reasoning data within \tool, we analyze the impact of two key factors on its performance: (1) the domain characteristics of the reasoning data, and (2) the selection of the model employed to generate the reasoning answer.

\subsubsection{Domain Characteristics}
Table~\ref{RQ4_type} investigates the impact of reasoning data domain characteristics on the effectiveness of vulnerability repair. We conduct the following variants: (1) using only the original dataset as a baseline (i.e., \textit{w/o Reason}), (2) augmenting the dataset with reasoning answers generated by RAC (i.e., \textit{w/ Reason}), (3) further enriching the data in (2) with mathematical domain samples from AMC~\cite{AMC} and AIME~\cite{AIME} (i.e., \textit{w/ Reason $\&$ Math}), and (4) further enriching the data in (2) with code domain samples from Codeforces~\cite{DBLP:journals/corr/abs-2501-01257/codeforce}  (i.e., \textit{w/ Reason $\&$ Code}).

As shown in Table~\ref{RQ4_type}, for both PrimeVul and SVEN, \tool achieves optimal performance when the reasoning data is constructed using RAC in combination with code domain data. This configuration yields improvements of 8.74\% and 6.88\%, respectively, in the EM metric.
In contrast, incorporating the training data with the mathematical domain does not yield additional performance gains. It leads to a reduction in CodeBLEU scores. 
We hypothesize that this is because code and vulnerabilities share greater formal and semantic similarity, whereas the addition of mathematical domain data may introduce redundancy, potentially hindering the model’s ability to effectively repair vulnerabilities.
\begin{table}[t]
\centering
\setlength{\tabcolsep}{2mm}
\renewcommand{\arraystretch}{1.2}
\caption{Impact of model selection for reasoning answer construction on \tool's performance.}
\resizebox{0.49\textwidth}{!}{
\begin{tabular}{l|cccccc}
\toprule

\rowcolor[HTML]{DEDEDE}
  & \multicolumn{3}{c}{\textbf{PrimeVul}~\cite{DBLP:journals/corr/abs-2403-18624/primevul}}  & \multicolumn{3}{c}{\textbf{SVEN}~\cite{DBLP:conf/ccs/HeV23/SVEN}}      \\
\rowcolor[HTML]{DEDEDE}
   \multirow{-2}{*}{\textbf{Variant}}                        & \textbf{Success $\uparrow$} & \textbf{EM$\uparrow$}   & \textbf{CodeBLEU $\uparrow$}  & \textbf{Success $\uparrow$} & \textbf{EM$\uparrow$}   & \textbf{CodeBLEU $\uparrow$}  \\

\midrule
\textit{- w/ Qwen2.5-14B}                                    & 75       & 17.24  & 39.20    & 106     & 28.80 & 54.41    \\
\textit{- w/ Qwen2.5-32B}                                    & 79       & 18.16  & 36.49    & 108     & 29.35 & 47.55    \\
\midrule
\rowcolor[HTML]{DEDEDE}
\textbf{\textit{- w/ OpenAI-o3}}        & \scimp{\textbf{97}}{$\uparrow$\textbf{18}} & \scimp{\textbf{22.30}}{$\uparrow$\textbf{4.14\%}}  & \scimp{\textbf{43.87}}{$\uparrow$\textbf{7.38\%}} & \scimp{\textbf{137}}{$\uparrow$\textbf{29}} & \scimp{\textbf{37.23}}{$\uparrow$\textbf{7.88\%}} & \scimp{\textbf{61.12}}{$\uparrow$\textbf{13.57\%}} \\
\bottomrule
\end{tabular}
}
\label{RQ4_model}

\end{table}
\subsubsection{Model Selection}
We evaluate the performance of \tool using reasoning answers generated by different models, including Qwen-14B-Instruct-1M~\cite{DBLP:journals/corr/abs-2501-15383/14B} (i.e., \textit{w/ Qwen2.5-14B}), Qwen-Coder-32B-Instruct~\cite{DBLP:journals/corr/abs-2409-12186/qwen2.5} (i.e., \textit{w/ Qwen2.5-32B}), and OpenAI-o3~\cite{gpto3}. The corresponding results are presented in Table~\ref{RQ4_model}. From these results, we observe that \tool achieves progressively better performance when utilizing reasoning data generated by more advanced models, with the best results obtained using OpenAI-o3. Furthermore, the quality of the reasoning data significantly impacts overall performance, which we attribute to the varying degrees of hallucination exhibited by the underlying models. Therefore, we recommend employing more powerful foundation models to generate reasoning answers for optimal results.

\begin{tcolorbox}[width=\linewidth,boxrule=0pt,top=1pt, bottom=1pt, left=1pt,right=1pt, colback=gray!20,colframe=gray!20]
\textbf{Answer to RQ4:} 
Optimal AVR performance is achieved when reasoning data is incorporated with the code domain, while incorporating mathematical domain data does not yield further improvements.
The effectiveness of reasoning data is highly dependent on the data quality, with stronger base models such as OpenAI-o3 producing superior results.
\end{tcolorbox}

\subsection{RQ5: Ablation Study}
\begin{table}[t]
\centering
\setlength{\tabcolsep}{2mm}
\renewcommand{\arraystretch}{1.2}
\caption{Ablation Study.}
\resizebox{0.49\textwidth}{!}{
\begin{tabular}{l|cccccc}
\toprule

\rowcolor[HTML]{DEDEDE}
  & \multicolumn{3}{c}{\textbf{PrimeVul}~\cite{DBLP:journals/corr/abs-2403-18624/primevul}}  & \multicolumn{3}{c}{\textbf{SVEN}~\cite{DBLP:conf/ccs/HeV23/SVEN}}      \\
\rowcolor[HTML]{DEDEDE}
   \multirow{-2}{*}{\textbf{Variant}}                       & \textbf{Success $\uparrow$} & \textbf{EM$\uparrow$}   & \textbf{CodeBLEU $\uparrow$}  & \textbf{Success $\uparrow$} & \textbf{EM$\uparrow$}   & \textbf{CodeBLEU $\uparrow$}  \\
\midrule
\textit{- w/o DAC} & \scdown{59}{$\downarrow$-49} & \scdown{13.56}{$\downarrow$-11.27\%} & \scdown{39.20}{$\downarrow$-6.97\%} & \scdown{112}{$\downarrow$-32} & \scdown{30.35}{$\downarrow$-8.78\%} & \scdown{56.68}{$\downarrow$-4.72\%} \\
\textit{- w/o \moduleB} & \scdown{97}{$\downarrow$-11} & \scdown{22.30}{$\downarrow$-2.53\%} & \scdown{43.87}{$\downarrow$-2.30\%} & \scdown{137}{$\downarrow$-7} & \scdown{37.23}{$\downarrow$-1.90\%} & \scdown{61.12}{$\downarrow$-0.28\%} \\

\midrule
\rowcolor[HTML]{DEDEDE}
\textbf{\tool} & \textbf{108} & \textbf{24.83} & \textbf{46.17} & \textbf{144} & \textbf{39.13} & \textbf{61.40} \\

\bottomrule
\end{tabular}
}
\vspace{-0.4cm}
\label{RQ5}

\end{table}

To assess the contributions of key components within \tool, we conduct a comprehensive ablation study, with the results summarized in Table~\ref{RQ5}. Specifically, we implement two variants of \tool: (1) one without the reasoning data generated by reasoning answer construction  (i.e., \textit{w/o RAC}), and (2) another without the curriculum-based verifiable rewarded training module (i.e.,  \textit{w/o \moduleB}), wherein only the SFT stage is retained. 

As shown in Table~\ref{RQ5}, the removal of either component leads to a noticeable degradation in performance. In particular, excluding the reasoning data results in an average performance drop of 10.03\% in EM and 5.85\% in CodeBLEU across both evaluated datasets. Similarly, omitting the \moduleB module yields a decline of 2.22\% in EM , highlighting the importance of curriculum-based reinforcement learning in enabling LLMs to learn from previously unexplored generations and enhance their generalization capability. 

\begin{figure}[t]
    \centering
    \begin{subfigure}[h]{0.24\textwidth}
        \centering
    	\includegraphics[width=1 \textwidth]{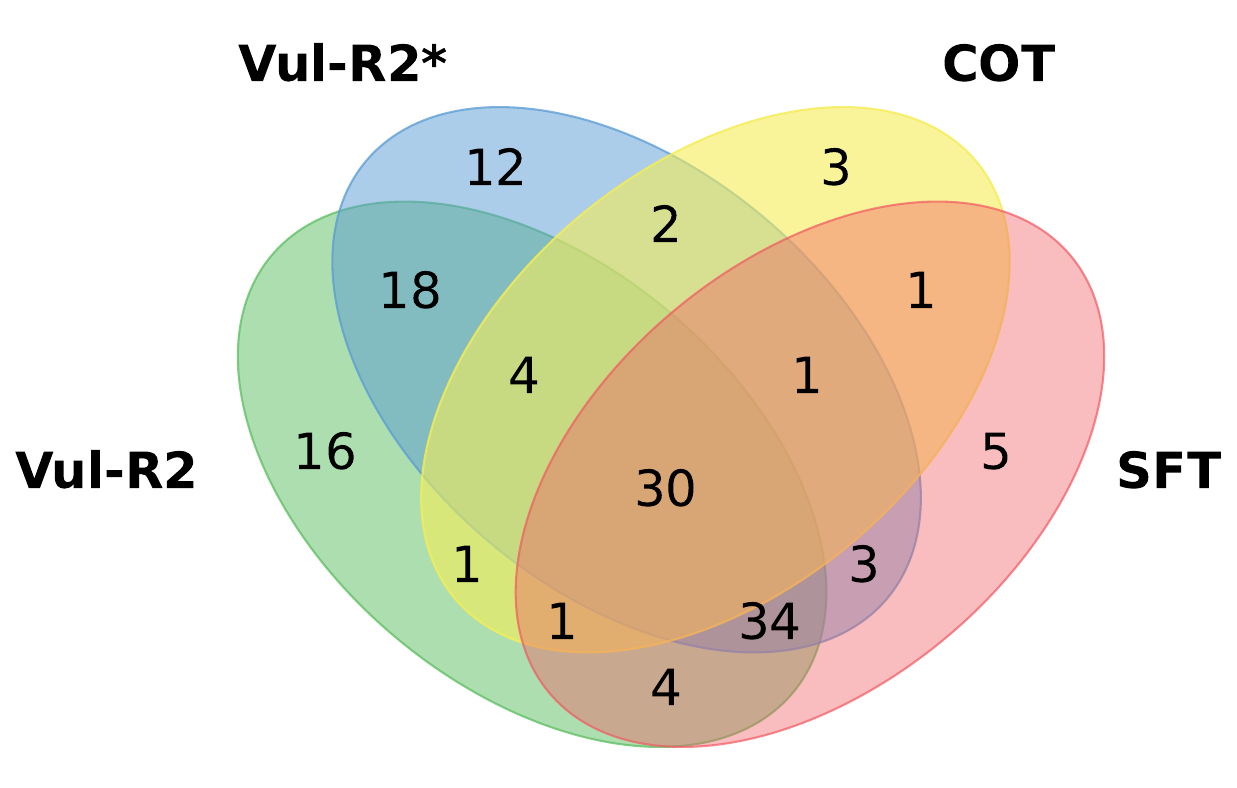}
        \vspace{-0.45cm}
    \caption{PrimeVul.}
    \end{subfigure}
    \hfill
    \begin{subfigure}[h]{0.24\textwidth}
        \centering
        \includegraphics[width=1 \textwidth]{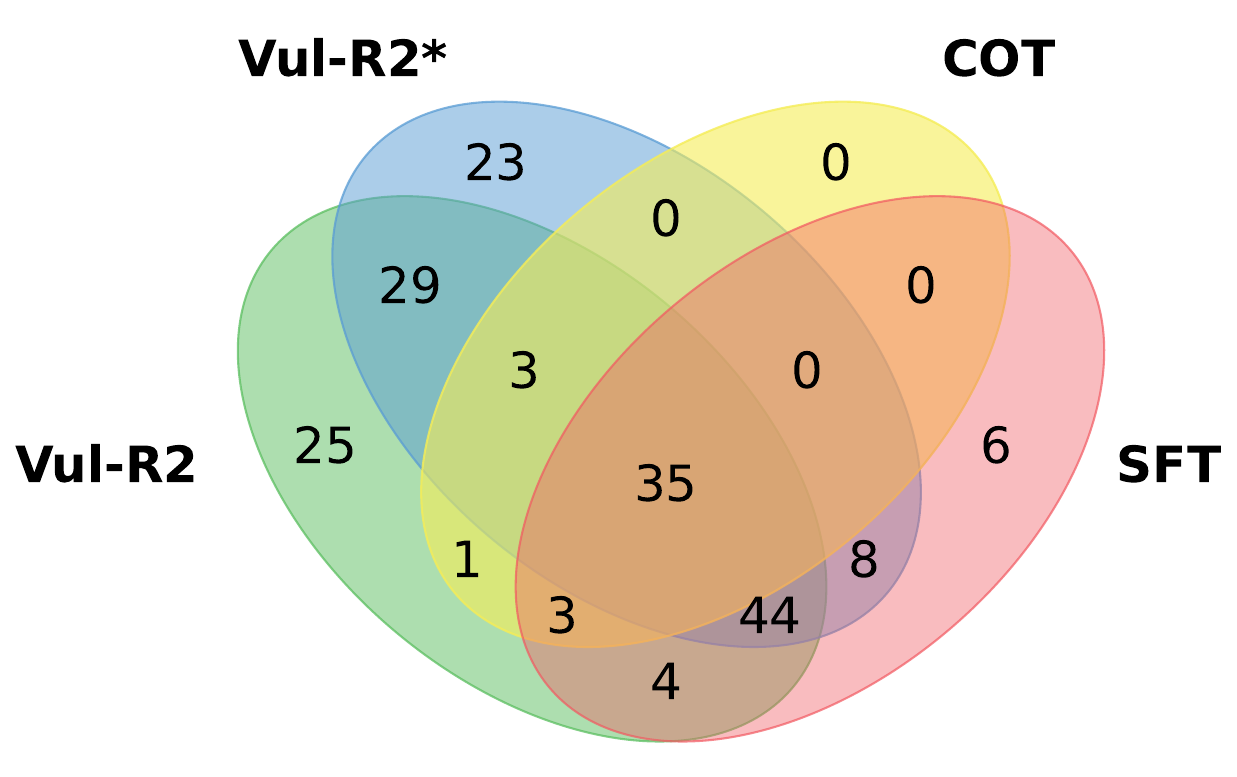}
        \vspace{-0.45cm}
        \caption{SVEN.}
    \end{subfigure}
    \caption{Venn diagram of the number of successfully fixed vulnerabilities for \textit{COT}, \textit{SFT}, $\text{\tool}^{*}$ (i.e., \textit{RLVR}), and \tool (i.e., \textit{SFT \& RLVR}) four paradigms.}
    \label{fig:venn_unique}
    \vspace{-0.45cm}
\end{figure}

In addition, we conduct comparative experiments on four paradigms: COT, SFT, $\text{\tool}^{*}$ (RLVR), and \tool (\textit{SFT \& RLVR}). As shown in Fig.~\ref{fig:venn_unique}, \tool\ and $\text{\tool}^{*}$ outperform the previous methods, successfully repairing 16 and 12 unique vulnerabilities on PrimeVul, respectively. 
This finding suggests that approaching the vulnerability repair task from a reasoning perspective through RLVR training can yield better results.

\begin{tcolorbox}[width=\linewidth,boxrule=0pt,top=1pt, bottom=1pt, left=1pt,right=1pt, colback=gray!20,colframe=gray!20]
\textbf{Answer to RQ5:} 
Each component of \tool plays a critical role in achieving optimal performance. 
Furthermore, we validate the effectiveness of the RLVR paradigm for AVR.
\end{tcolorbox}
\begin{figure}
    \centering
    \includegraphics[width=0.48\textwidth]{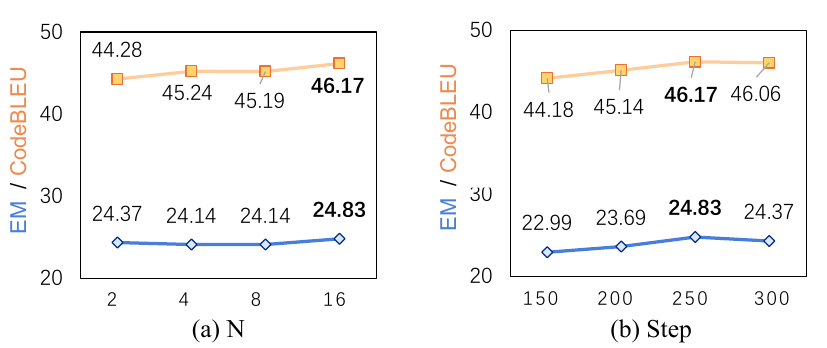}
    \caption{Parameter analysis.}
    \label{fig:parameter}
    \vspace{-0.4cm}
\end{figure}

\subsection{RQ6: Parameter Analysis}
In this section, we investigate the influence of two parameters: the number of generated examples ($N$) in RLVR and the training step, on the performance of \tool in the PrimeVul. 

\paragraph{Number of Generated Examples by RLVR} We study the effect of $N$ by varying it from 2 to 16. As depicted in Fig.~\ref{fig:parameter} {(a)}, \tool exhibits optimal performance when $N$ is set to 16, with larger values yielding improved results. We suggest that this is because selecting more samples increases the exploration space, making it easier to obtain correct rewards in RLVR. However, increased exploration also leads to higher computational overhead.

\paragraph{Number of Training Steps} We evaluate the performance of \tool across different training steps in RLVR. The results are presented in Fig.~\ref{fig:parameter} {(b)}. We find that \tool achieves the best performance at 250 steps, as further increasing the number of training steps may introduce the risk of overfitting problems due to the limited data. Consequently, we set the maximum number of iterations to 250 steps.

\begin{tcolorbox}[width=\linewidth,boxrule=0pt,top=1pt, bottom=1pt, left=1pt,right=1pt, colback=gray!20,colframe=gray!20]
\textbf{Answer to RQ6:} 
The performance of \tool improves as more samples are generated;  however, large numbers lead to increased computational resource consumption. \tool achieves its best performance with 250 training steps.
\end{tcolorbox}
\section{Discussion}\label{sec:discussion}

\begin{figure*}[t]
	\centering
	\includegraphics[width=0.9\textwidth]{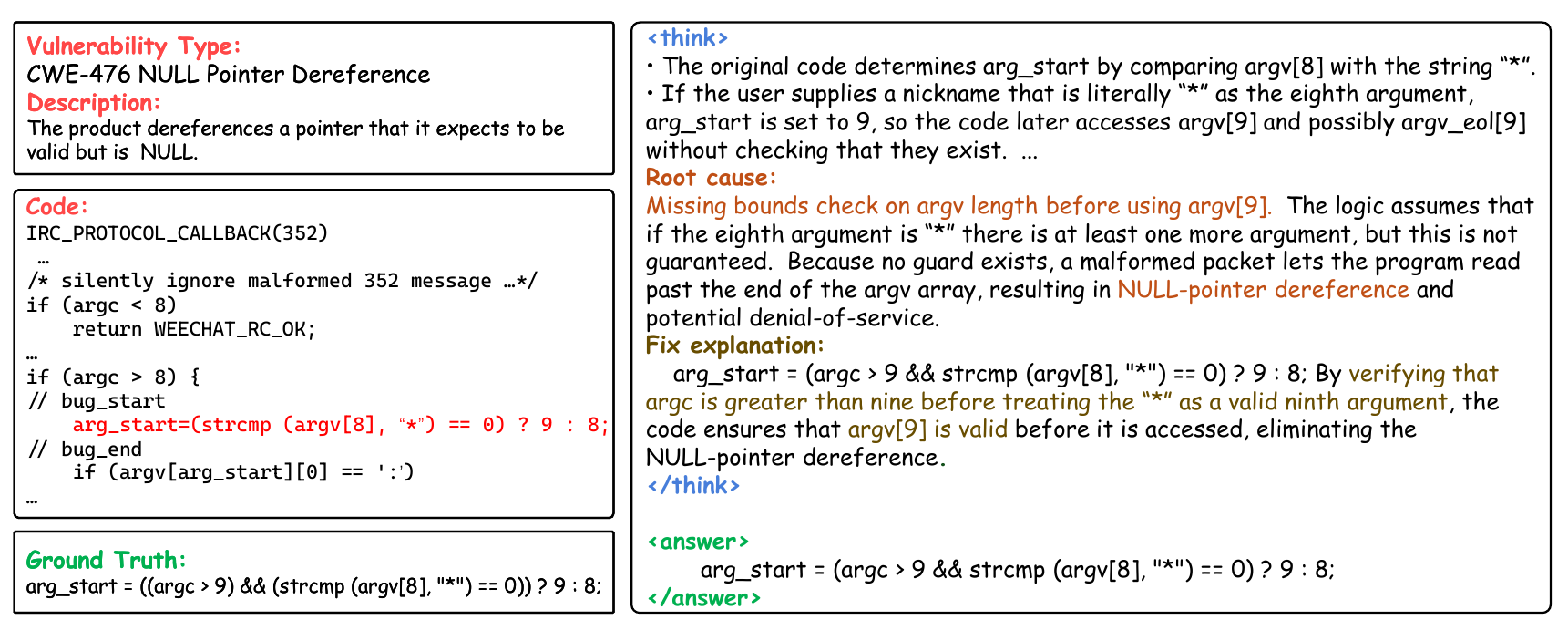}
    \caption{\textbf{Case Study}. A case of CVE-2020-9759~\cite{CVE-2020-9759} showing that \tool can accurately repair the vulnerability with detailed, interpretable reasoning answer, as illustrated in the right-side figure.} 
    \label{fig:detailed_case}
    \vspace{-0.4cm}
\label{example}
\end{figure*}

\subsection{What makes \tool work?}

\paragraph{Empowering LLMs with vulnerability-specific knowledge for AVR}
\tool can effectively equip LLMs with explicit knowledge of vulnerability patterns. As illustrated in Fig.~\ref{example}, this example demonstrates a case of ``NULL Pointer Dereference'' where all baseline methods failed to correctly fix the vulnerability, whereas \tool successfully repairs it.
LLMs primarily rely on general knowledge acquired during pre-training. Consequently, without explicit training on vulnerabilities, LLMs often struggle to repair vulnerabilities in real-world scenarios.
In contrast, \tool accurately captures the vulnerability pattern in this case, the code fails to check whether the number of parameters is sufficient, which may lead to out-of-bounds access or dereferencing a NULL pointer. By explicitly modeling and leveraging vulnerability patterns, rather than focusing solely on code semantics, \tool demonstrates superior effectiveness.

\paragraph{
More detailed reasoning process by \moduleB empowers LLMs for more effective AVR}
The second advantage of \tool lies in its RLVR-based \moduleB module, which enhances the reasoning capabilities of LLMs by dynamically constructing verifiable reward signals into the training process. In contrast, prior works~\cite{DBLP:conf/icse/ZhouKXH024/vulmaster, DBLP:conf/sigsoft/FuTLNP22/vulrepair} typically generate vulnerability patches directly, without a reasoning process. As illustrated in Fig.~\ref{example}, \tool first analyzes the original code to identify existing issues. Then, it pinpoints the root cause of a vulnerability, such as the ``missing bounds check on argv length before using argv[9]'' (mentioned in ``Root cause'').  Subsequently, \tool generates a detailed fix explanation, which includes ``verifying'' the proposed solution and the steps taken to ``ensure'' its correctness (mentioned in ``Fix explanation'').
\tool systematically trains LLMs through each stage of reasoning, 
with the process being guided by verifiable feedback.
Our results demonstrate that the strong reasoning capabilities enabled by \moduleB improve vulnerability repair.

\subsection{Threats to Validity}
We have identified the following threats and limitations:
\paragraph{Generalizability to Other Programming Languages}
In this work, we conduct experiments on the PrimeVul and SVEN datasets, which cover two widely-used programming languages: C and C++. Although additional datasets exist for other languages, such as the Java dataset in the Vul4J~\cite{DBLP:conf/msr/BuiSF22/Vul4J} Benchmark. We do not include them in our evaluation due to an insufficient number of samples to support the training objectives of RLVR. In future work, we plan to continue collecting data and to extend our experiments to encompass a broader range of programming languages.

\paragraph{Constraints on Reasoning Context Length}
All reasoning samples in our experiments are less than 4,096 tokens in length. Consequently, \tool may encounter difficulties in repairing vulnerabilities within code snippets that exceed this length, potentially resulting in the loss of relevant contextual information due to truncation. This limitation may affect the evaluation of longer code segments in the context of software vulnerability repair. We plan to address this issue in future work by conducting experiments with increased computational resources to accommodate longer input sequences.

\paragraph{Limitations in LLM Selection}
Another potential threat to validity stems from the selection of foundation models used in \tool. Following prior work~\cite{DBLP:journals/corr/abs-2502-14768/logicrl}, we evaluate \tool using the Qwen-2.5 series~\cite{DBLP:journals/corr/abs-2409-12186/qwen2.5}. In future work, we intend to further assess the effectiveness of \tool across a wider variety of LLMs.

\section{Related Work}\label{sec:related}
\subsection{Automated Vulnerability Repair}
Previous research has proposed various methods for AVR, which can be broadly categorized into supervised learning-, CodePTM-, and LLM-based methods. 
Supervised learning-based approaches~\cite{DBLP:journals/tse/ChiQLZY23/Seqtrans, DBLP:conf/esorics/MaTLSD17/VuRLE, DBLP:conf/sigsoft/FuTLNP22/vulrepair, DBLP:conf/icse/ZhouKXH024/vulmaster, DBLP:conf/aiware/KulsumZXd24/VRPilot} learn to generate patches by observing patterns in labeled vulnerability-fix pairs. For example, Vurle\cite{DBLP:conf/esorics/MaTLSD17/VuRLE} represents one of the earliest learning-based frameworks, which learns contextual code transformations directly from examples of vulnerable code and their corresponding fixes. CodePTMs are neural models pretrained on large-scale corpora of code and are often fine-tuned on curated datasets of vulnerability fixes. For example, VulRepair\cite{DBLP:conf/sigsoft/FuTLNP22/vulrepair} fine-tunes CodeT5 using a byte pair encoding tokenizer~\cite{BPE} and the CVEFixes dataset \cite{CVEFixes}. Similarly, VulMaster\cite{DBLP:conf/icse/ZhouKXH024/vulmaster} also builds upon CodeT5, augmenting the model with abstract syntax trees and CWE examples generated by ChatGPT \cite{ChatGPT}.
LLM-based approaches~\cite{DBLP:journals/corr/abs-2107-03374/Codex, DBLP:journals/corr/abs-2409-12186/qwen2.5, DBLP:journals/corr/abs-2407-21783/llama3.1} use zero-shot or few-shot prompting techniques to AVR. 
For example, Wu et al. \cite{DBLP:conf/issta/WuJPLD0BS23/wu} conducted evaluations of LLMs on the Vul4J dataset \cite{DBLP:conf/msr/BuiSF22/Vul4J}.

\subsection{RLVR and Software Engineering}
RLVR has recently emerged as a promising method for enhancing the reasoning capabilities of LLMs~\cite{DBLP:journals/corr/abs-2503-15478/rl1, DBLP:journals/corr/abs-2502-14768/logicrl, DBLP:journals/corr/abs-2501-12948/DeepSeekR1, DBLP:journals/corr/abs-2502-18449/swerl, DBLP:journals/corr/abs-2504-14286/srpo}. For instance, DeepSeek-AI introduced GRPO and DeepSeek-R1~\cite{DBLP:journals/corr/abs-2501-12948/DeepSeekR1}, showcasing the effectiveness of RLVR in improving LLM reasoning performance. SWE-RL~\cite{DBLP:journals/corr/abs-2502-18449/swerl} pioneered the application of reinforcement learning for LLM-based reasoning in real-world software development tasks. Furthermore, SRPO~\cite{DBLP:journals/corr/abs-2504-14286/srpo} proposed a cross-domain training framework that jointly optimizes mathematical reasoning and programming proficiency, demonstrating improved generalization across diverse task categories.  Despite these advancements, 
our approach is the first to train LLMs by RLVR in the vulnerability domain, aiming to enhance the reasoning capabilities of LLMs without relying on intermediate feedback from the environment. 


\section{Conclusion and Future Work}\label{sec:conclusion}
In this paper, we propose \textit{Vulnerability Reasoner and Repair (\tool)}, a reasoning LLM for vulnerability repair. It models the vulnerability repair task from a reasoning perspective, which comprises two key components: a domain-aware reasoning learning module and a curriculum-based verifiable rewarded training module. \tool enables the model to reason about complex vulnerability patterns and generate effective patch candidates for AVR.
We conduct comprehensive experiments on the PrimeVul and SVEN datasets to evaluate the effectiveness of \tool in enhancing vulnerability repair capabilities. 
In the future, we plan to apply our reasoning LLM to other tasks in the vulnerability domain, such as
vulnerability detection.
Our source code is available at \http.

\section*{Acknowledgment}
This research is supported by the National Natural Science Foundation of China under project (No. 62472126, 62276075), Natural Science Foundation of Guangdong Province (Project No. 2023A1515011959), and Shenzhen-Hong Kong Jointly Funded Project (Category A, No. SGDX20230116 091246007).





\bibliographystyle{IEEEtran}
\bibliography{custom}

\end{document}